# Societal attitudes toward service robots: Adore, abhor, ignore or unsure?

Yoganathan, V., Osburg, V.-S., Fronzetti Colladon, A., Charles, V., Toporowski, W.

**Abstract**

Societal or population-level attitudes are aggregated patterns of different individual attitudes, representing collective general predispositions. As service robots become ubiquitous, understanding attitudes towards them at the population (vs. individual) level enables firms to expand robot services to a broad (vs. niche) market. Targeting population-level attitudes would benefit service firms because: 1) they are more persistent, thus, stronger predictors of behavioral patterns, and 2) this approach is less reliant on personal data, whereas individualized services are vulnerable to AI-related privacy risks. As for service theory, ignoring broad unobserved differences in attitudes produces biased conclusions, and our systematic review of previous research highlights a poor understanding of potential heterogeneity in attitudes toward service robots. We present five diverse studies (S1-S5), utilizing multinational and 'real world' data ($N_{total}$ = 89,541; years: 2012-2024). Results reveal a stable structure comprising four distinct attitude profiles (S1-S5): *positive* ("adore"), *negative* ("abhor"), *indifferent* ("ignore"), and *ambivalent* ("unsure"). The psychological need for interacting with service staff, and for autonomy and relatedness in technology use, function as attitude profile antecedents (S2). Importantly, the attitude profiles predict differences in post-interaction discomfort and anxiety (S3), satisfaction ratings and service evaluations (S4), and perceived sociability and uncanniness based on a robot's humanlikeness (S5).

# INTRODUCTION

In an era of technologically augmented services, service robots have emerged as an integral feature of services in multiple industries. Service robots (as opposed to industrial robots) are "system-based autonomous and adaptable interfaces that interact, communicate, and deliver services to an organization's customers" (Wirtz et al. 2018 p. 907). The global service robot market is estimated to be worth $28.9 billion in 2024, more than three times that of the industrial robot market, and is predicted to grow to $34.7 billion in 2028 (Statista 2024). With such dramatic expansion in service robot implementation anticipated across industries, there is greater need for firms to understand attitudes toward service robots at a broader (i.e., societal/population) level, not least because prior attitudes influence customer evaluations and subsequent use of robot-delivered services (Crolic et al. 2022). While scholars have alluded to substantial heterogeneity in attitudes toward service robots (e.g., Mende et al. 2019), previous research has given scant attention to understanding the characteristics of this heterogeneity, especially at the population (as opposed to individual) level. Unlike observable customer differences (i.e., demographics), attitudinal heterogeneity is unobserved (or latent). Therefore, understanding the characteristics of attitudinal heterogeneity represents an important challenge for practitioners and theorists alike. Consequently, our investigation focuses on ascertaining 1) the structure of attitudes toward robots at the population level, 2) the stability of that structure over time, 3) its antecedents, and 4) its consequences in terms of customer evaluations. To this end, we present a diverse empirical package to maximize rigor and impact, comprising data sources from multiple countries spanning a 12-year period.

By *attitude toward service robot*s, we refer to an overall evaluation or predisposition pertaining to service robots (Stapels and Eyssel 2021). Aggregated patterns of different individual attitudes are described as *population-level attitudes*, which represent collective predispositions among the general population (Charlesworth and Banaji 2021). While *attitude structure* refers to a typology of attitudes towards an entity (Poortinga and Pidgeon 2006), *attitude stability* describes the persistence of attitudes over time (Glasman and Albarracín 2006).

Discerning stable (vs. transient) attitudes is pivotal to the efficacy of target marketing strategies, owing to their predictability and stronger influence on behavior (Glasman and Albarracín 2006). Yet, previous studies on service robots have limited their focus to individual (vs. population-level) attitudes, which alter based on individual experiences or learning over time. On the other hand, population-level attitudes are much more persistent, thus exerting a stronger impact on behavioral patterns (Charlesworth and Banaji 2021; Glasman and Albarracín 2006). The dynamic developments in service robot technology are likely to exacerbate the variability in corresponding individual attitudes (Greengard 2024), which makes these impractical for firms to target. Furthermore, privacy and data risks associated with Artificial Intelligence (AI) can create barriers to highly individualized services, as these require firms to leverage personal customer data (Puntoni et al. 2021). Therefore, as the expansion in the deployment of service robots accelerates, understanding population-level attitudes is imperative to service firms.

From a service theory perspective, despite predictions based on attitude theory for the existence of multiple types of attitude (Pidgeon et al. 2005), current literature on service robots only considers negative and/or positive attitudes. This simplistic approach overlooks the fact that some people may be apathetic or have mixed feelings about service robots (Stapels and Eyssel 2021). Inadequately accounting for attitudinal heterogeneity in this way can bias results and impede researchers from developing an accurate or more complete understanding of the impact of service robots on customers (see: Becker et al. 2013; Bryan, Tipton, and Yeager 2021).

In this context, our investigation utilizes multiple methods and data sources from different countries and time-periods to provide a truer representation and a comprehensive understanding of population-level attitudes toward service robots.

## EMPIRICAL OVERVIEW

To address the gaps discussed, we present five studies (see Table 1), comprising 89,541 total individual records collected in 2012-2024 from multiple countries. Study 1 utilizes a large archival dataset collected in three waves from 29 European countries, which enables a generalizable investigation into population-level attitudes (Gnambs and Appel 2019). We gather primary data in Study 2 to replicate the identified attitude structure and ascertain its theory-based antecedents. Study 3 examines participants' actual interactions with a state-of-the-art virtual service robot to reveal between-profile differences in post-interaction discomfort and anxiety. In Study 4, we conduct psycholinguistic analyses on textual online reviews of internationally renowned robot hotels, elucidating attitude profile differences in service evaluations and satisfaction ratings. Finally, Study 5 leverages secondary data to identify within- and between-profile differences in how robots' humanlike design affects socio-cognitive evaluations (i.e., perceived sociability, agency, and uncanniness). Thus, by combining primary and secondary sources, and incorporating 'real world' data, we enhance our findings' validity, generalizability, and ecological value (van Heerde et al. 2021).

**Table 1. Overview of Studies**

| | Study 1 | Study 2 | Study 3 | Study 4 | Study 5 |
|---|---|---|---|---|---|
| **Study Aims** | Identify structure and test stability of attitudes at population level. | 1) Replicate the identified attitude structure.<br>2) Examine theory-based antecedents of attitude profiles. | 1) Examine differences between attitude profiles in terms of discomfort and anxiety in interacting with service robots based on actual interaction.<br>2) Validate attitude profiles and add ecological value. | 1) Examine differences between attitude profiles in terms of customer satisfaction ratings based on online reviews of robot hotels.<br>2) Validate attitude profiles and add ecological value. | Examine anthropomorphic effects of attitude profiles *vis-à-vis* a robot's degree of humanlike design. |
| **Key Findings** | Attitude structure of four profiles – negative, indifferent, ambivalent, and positive – is stable over time (and cultures in Europe). | Need for interaction with service employees, and basic psychological needs in technology use predict attitude profile membership. | Indifferent attitude predicts the second lowest level of post-interaction discomfort and anxiety, while ambivalent attitude exhibits the second highest levels of discomfort and anxiety. Negative and positive attitudes reveal the highest and lowest levels respectively. | Indifferent attitude predicts the second highest level of customer satisfaction rating, while ambivalent attitude reveals the second lowest level of customer rating. The emotionality, certainty, extremity, and valence of reviews help identify the attitudes profiles. | Average sociability for indifferent is lower than that of ambivalent, but indifferent also has a lower tendency for uncanniness. 'Nimbro' elicits the most positive overall evaluations in all profiles. |
| **Design** | Three-wave multinational survey (2012-2017). | Online survey (2024). | Interaction with virtual robot; online survey (2024). | English language online reviews of robot hotels (2022). | Online survey (2021). |
| **Source and Sample Size** | Secondary: European Commission (n=82,453) | Primary: Prolific (n=950) | Primary: Prolific (n=301) | Secondary: Google, TripAdvisor, and Booking.com. (n=4,696) | Secondary: Spatola and Wykowska (2021) (n=1,141). |
| **Analytical Methods** | Latent profile models incorporating predictors and groups.<br>Double Machine Learning treatment effects model. | Latent profile model with predictors. | Latent profile model incorporating a path model.<br>Nonparametric repeated measures analysis. | Psycholinguistic analysis of textual data.<br>Finite mixture path model.<br>Qualitative text analysis. | Latent profile model incorporating a path model. |

We reveal four distinct and stable attitude profiles (S1-S5): *positive* (low risk and high benefit perceptions), *negative* (high risk and low benefit perceptions), *indifferent* (low risk and low benefit perceptions), and *ambivalent* (high risk and high benefit perceptions). Further, the need for relatedness and autonomy in technology use, and the need for interaction with service staff, act as attitude profile antecedents (S2). Notably, the attitude profiles predict differences in post-interaction discomfort and anxiety (S3), satisfaction ratings and service evaluations of robot hotels (S4), and perceived sociability and uncanniness based on a robot's humanlikeness (S5).

Our findings entail several implications for service firms. For the negative profile, firms should emphasize human interaction through staff availability and fostering customer socialization, and support customers' autonomy by enabling them to choose between human and robot service alternatives. For the ambivalent profile, firms should balance robot technology with conventional amenities that add value, and improve the reliability of robot services, while ensuring human staff are available. Firms should focus on creating memorable experiences involving robots for the indifferent profile, and avoid overemphasizing robot services. For the positive profile, while leveraging its enthusiasm for publicizing robot services (e.g., via social media), continuous customer engagement through enhanced robot services and loyalty programs is recommended. Robots with median humanlikeness (e.g., Nimbro) are best suited to appeal to the broadest range of customer attitudes based on perceived sociability, agency, and uncanniness.

## THEORETICAL BACKGROUND AND RESEARCH GAPS

Despite their increasing prevalence in service frontlines, service robots are still a novel form of innovation, which significantly differ from predecessors like smartphones, computers, or self-service technologies (McLeay et al. 2021). For instance, they have the ability to decide and act independently (i.e., autonomy). Such humanlike aspects of service robots can induce nuanced and persuasive effects on customers (Wirtz et al. 2018). Therefore, attitudes toward service robots warrant a more specialized understanding. To this end, we systematically reviewed previous empirical studies related to attitudes toward service robots (procedure explained in Web Appendix A). Consequently, we identified five research gaps (see tabulated summary in Web Appendix B).

First (Gap 1), although there is now a burgeoning body of research focusing on service robots, the structure of attitudes towards them is poorly understood. In our reading, the majority of articles consider attitudes to be bipolar (negative-positive) and do not account for the underlying complexity in terms of the potential for several different types of attitudes when estimating focal effects (e.g., Akdim, Belanche, and Flavián 2023; Guan et al. 2022; Han, Deng, and Fan 2023). Nascent evidence shows that the bipolar conceptualization is flawed in that it omits consideration of attitudinal ambivalence. For example, Dang and Liu (2021) found that robots with high (vs. low) mental abilities induce more ambivalent attitudes, which are distinct from negative and positive attitudes. Social robotics literature reveals that the structure of attitudes toward robots is multidimensional, rather than a bipolar continuum encompassing positive, negative, or even neutral attitudes (Stapels and Eyssel 2021). Thus, multiple attitude profiles may exist, but more in-depth empirical understanding is needed.

Second (Gap 2), assessing the stability of attitude profiles is important, since stable attitudes exert greater influence on behavioral outcomes (Glasman and Albarracín 2006). Further, if overlooked in empirical research, attitudinal heterogeneity that remains unchanged over time may give rise to an inaccurate or incomplete understanding of service robots' impact on customers (Becker et al. 2013). The following set of observations illustrates this. While some studies find favorable attitudes toward service robots (e.g., Gelbrich, Hagel, and Orsingher 2021; Yam et al. 2021), others point to a negative disposition (e.g., Castelo et al. 2023; Luo et al. 2019). Such inconsistencies can be attributed mainly to the failure to account for persistent population-level attitude heterogeneity (Osburg et al. 2022). Moreover, hardly any studies have explicitly examined the stability of attitude profiles (see tabulated overview in Web Appendix B). This distorted empirical landscape also prevents the development of effective interventions to avert or mitigate undesirable outcomes and foster beneficial ones (Bryan et al. 2021). Impediments to effective interventions are even more problematic given the stronger influence of stable attitudes on behavior. Service providers too would benefit from understanding stable (vs. changeable) attitudes, since stable attitudes can be reliably targeted. Therefore, ascertaining attitude stability is as important as identifying an attitude structure.

Third (Gap 3), there are important reasons for focusing on attitudes at the population (vs. individual) level. When expanding robot services to a broad (vs. niche) market, it is vital that firms understand population-level attitude structure and stability, so as to address potential resistance more effectively. The predictability of attitudes is also important for attitude-specific targeting of service strategies. However, individual attitudes can vary easily (e.g., based on direct experience) and have relatively limited generalizability – e.g., elderly customers' attitudes toward a care service robot improved over time through usage (Stafford et al. 2014), and the COVID-19 pandemic may have had a similar effect (Odekerken-Schröder et al. 2020). In contrast, attitudes are much more stable at the population level (Charlesworth and Banaji 2021), whereby individual attitude variations are subsumed into population-level distributions (or "attitude space") (Glasman and Albarracín 2006; Poortinga and Pidgeon 2006). Moreover, since personalized robot services (i.e., reliant on personal data) are vulnerable to privacy risks (Čaić, Odekerken-Schröder, and Mahr 2018; Puntoni et al. 2021), targeting population-level attitudes can be a viable alternative for firms, as this approach is substantially less reliant on individual customer data. Our review illustrates that, except for Gnambs and Appel (2019) who used population-level representative data, only Binesh and Baloglu (2023) examined attitudes at an aggregate level (positive and negative clusters).

Fourth (Gap 4), we consider multi-method or multi-source data important in investigating potentially stable attitude structures at the population level. This is because multiple sources of evidence can help validate findings and generalize results to the population (e.g., Castelo et al. 2023; Crolic et al. 2022). For example, if multiple samples from different time periods or contexts reveal the same pattern of results, this would provide robust evidence of stability. While our review identified several studies with multi-source data, most of these are online experiments, conducted within relatively short time frames. Finally (Gap 5), as per the theory of anthropomorphism, customer evaluations and intentions are influenced by a robot's design – e.g., humanlike embodiment (Blut et al. 2021; Spatola and Wykowska 2021). These influences can be quite nuanced; for example, Kim et al. (2019) find that anthropomorphic robots elicit greater warmth perceptions, but ultimately exert a negative impact on attitude due to the uncanniness caused by anthropomorphism. Therefore, it is necessary to understand how such nuanced evaluations are affected by the degree of robots' humanlike design in light of different attitudes toward service robots. A related limitation is that only a handful of studies are based on actual interactions with robots (e.g., Mende et al. 2019), which raises questions about the ecological value of findings from current literature.

## HYPOTHESIS DEVELOPMENT

### Attitude Structure and Stability

Initial theorization of attitude structure was based on a conceptualization of attitudes as operating on a bipolar dimension of negative vs. positive valence, and attitudes that were neither positive nor negative were assumed to be neutral (Cacioppo, Gardner, and Berntson 1997). This theorization, as pointed out earlier, has continued to influence the extant literature on attitudes towards service robots. In contrast, based on the simultaneous espousal of both positive and negative perceptions (e.g., perceiving both risks and benefits of a product/service), attitude theorists have long since advanced a more expanded structure of attitudes – i.e., a 2 (benefits: high/low) x 2 (risks: high/low) typology of attitudes (Pidgeon et al. 2005). Notably, this typology treats ambivalence and indifference as distinct attitudes, rather than as "middle of the road" or neutral positions on a positive-negative continuum (*ibid*). This theorization also resonates with more recent findings of social robotics research in terms of ambivalent attitudes towards robots (Stapels and Eyssel 2021). Furthermore, Poortinga and Pidgeon's (2006) empirical examination of attitude structure using the risk-benefit typology revealed that attitude scores manifest along three underlying dimensions – *positive vs. negative* (positively and negatively valanced attitudes), *involved vs. uninvolved* (indifferent attitude), and *certain vs. uncertain* (ambivalent attitude). These dimensions give rise to an *attitude space* (see Figure 1), through which individuals may move as their perceptions of benefits and risks change over time (*ibid*). This theory of attitude structure has been applied to understanding various customer attitude types. For example, Liu and Xu (2020) examined the public's attitude toward autonomous transport and identified the same four-profile structure as Poortinga and Pidgeon (2006). Moreover, despite some changes to individual attitudes through direct experience, Liu and Xu (2020) find four stable attitude profiles – i.e., the structure conforms with attitude theory. Thus, although individual attitudes can change over time through learning and experience, the attitude structure can remain consistent at population-level (Charlesworth and Banaji 2021).

**H1**: The structure of attitudes toward service robots at the population level comprises *positive* (low risk, high benefit), *negative* (high risk, low benefit), *indifferent* (low risk, low benefit), and *ambivalent* (high risk, high benefit) attitude profiles.

**H2**: The attitude structure in H1 is stable over time, such that the four profiles persist.

**Figure 1. Conceptual Model of Attitude Structure**

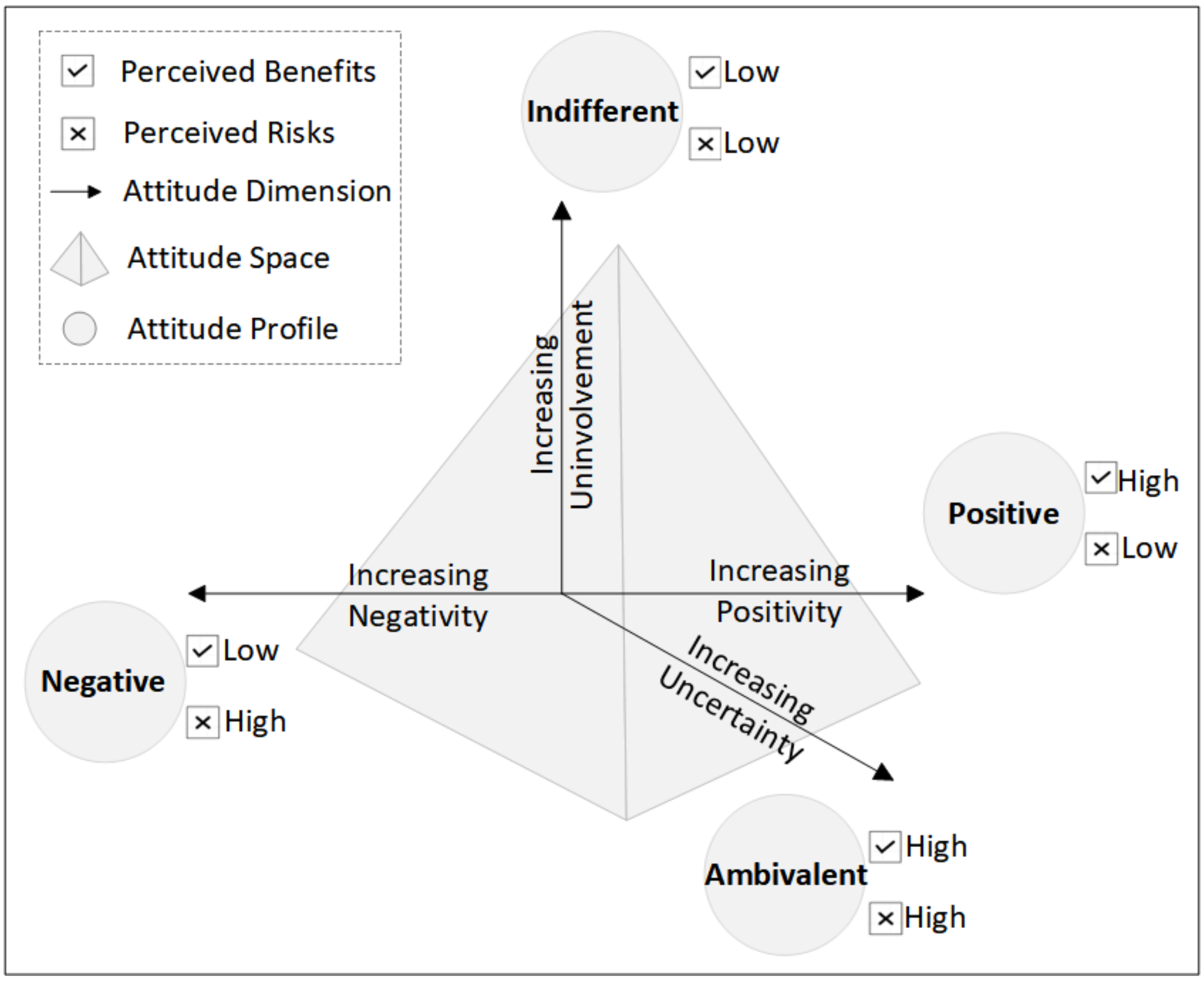

*Note*. Developed by authors by adapting the work of Poortinga and Pidgeon (2006).

## Psychological Antecedents of Attitude Profiles

The attitude structure discussed earlier is based on the psychological assessments of risks and benefits. Such risk and benefit assessments, particularly in the technology adoption context, are shaped by underlying human motivations for need fulfillment (Ehrari, Ulrich, and Andersen 2020). The need for autonomy (to be in control and have freedom), competence (to be effective and have mastery), and relatedness (to feel connected, cared for and able to care) are recognized as fundamental human motivations, the fulfillment of which drives human behavior, including technology adoption (Ryan and Deci 2000). Research on broad AI-related attitudes has revealed that the fulfillment of autonomy, competence, and relatedness through technology decreases negative attitudes toward AI (Bergdahl et al. 2023). Thus, the fulfillment of autonomy, competence, and relatedness are critical to customer wellbeing and reduce service robot adoption barriers (Janssen and Schadenberg 2024).

Conversely, when basic psychological needs are unmet when using technology, individuals are likely to develop resistance. For example, those who have a need for competence (i.e., no competence fulfillment) in technology use may not be self-confident in managing risks, or exploiting benefits, associated with robot-delivered services (Bergdahl et al. 2023). Service robots' autonomous and intelligent capabilities can also increase risk (and reduce benefit) perceptions by threatening human identity and suppressing feelings of autonomy in customers (Odekerken-Schröder et al. 2020). Consequently, the ability to control service robots enables customers to react more positively towards them, e.g., by assuming responsibility for negative service outcomes involving robots (Jörling, Böhm, and Paluch 2019). Furthermore, service robots may exert an isolating influence on customers by substituting interpersonal connections with impersonal service provision, such that individuals who have a need for relatedness in technology use may adopt a stronger resistance to service robots (Čaić et al. 2018; Odekerken-Schröder et al. 2020). An alternative form of the need for satisfying relatedness could be that customers adopt a psychological tendency to seek interactions with service employees and adopt an aversion to interacting with machines instead of service employees (Dabholkar 1996). Prior research on service robots has revealed that the need for interaction with service employees is a strong predictor of psychological risk associated with a service (Yoganathan et al. 2021). Based on the preceding arguments, we anticipate that the need for autonomy, competence, and relatedness in the use of technology, as well as the need for interaction with service employees will increase the likelihood of resistant attitudes towards service robots.

**H3**: The need for autonomy in the use of technology increases the likelihood of *negative* (H3a), *ambivalent* (H3b), and *indifferent* (H3c) attitude profiles compared to the *positive* attitude profile.

**H4**: The need for competence in the use of technology increases the likelihood of *negative* (H4a), *ambivalent* (H4b), and *indifferent* (H4c) attitude profiles compared to the *positive* attitude profile.

**H5**: The need for relatedness in the use of technology increases the likelihood of *negative* (H5a), *ambivalent* (H5b), and *indifferent* (H5c) attitude profiles compared to the *positive* attitude profile.

**H6:** The need for interacting with service employees increases the likelihood of *negative* (H6a), *ambivalent* (H6b), and *indifferent* (H6c) attitude profiles compared to the *positive* attitude profile.

## STUDIES 1 AND 2: ATTITUDE STRUCTURE, STABILITY AND ANTECEDENTS

### Study 1: Data and Measures

We drew upon archival data from the European Commission collected in three separate waves (2014, 2017, 2018) from 29 European countries. The full sample comprises 82,453 individual responses ($n_{2012}$ = 26,751; $n_{2014}$ = 27,801; $n_{2017}$ = 27,901), which we utilized in its entirety. The data were collected by adopting a stratified random sampling process. Each respondent was interviewed in their native language, and their responses were recorded by the interviewers following a standardized approach. Thus, the dataset is representative of the adult population in each sampled country and a reliable source of public opinion (Gnambs and Appel 2019). To assess attitude structure, we identified four attitudinal questions (see Web Appendix C for descriptives), which relate to perceived benefits ("*Robots help people*" and "*Robots do jobs that are hard/dangerous for people*") and perceived risks ("*Robots steal jobs*" and "*Robots require careful management*"). These variables are consistent with attitude structure theory discussed earlier (Pidgeon et al. 2005; Poortinga and Pidgeon 2006), and empirical research focusing on social robots and autonomous vehicles (Liu and Xu 2020; Stapels and Eyssel 2021). We further identified several demographic and technology-related factors as potential predictors of profile membership: *age*, *age at full-time education ceased*, *sex*, *occupation*, *frequency of internet usage*, and *prior experience with robots*.

**Study 1: Analysis and Results**

*Model Specification*

We deployed a finite mixture approach to identifying the attitude structure using a latent profile analysis, which is a well-established model-based method for revealing the natural typology (as opposed to a forced classification) of unobserved heterogeneity (Osburg et al. 2022). Adopting this approach allows us to identify attitude distributions (or profiles) that may exist in the population at an aggregate (as opposed to individual) level, whereby each individual (observation) is accorded a probability of exhibiting each of the resulting attitude profiles (Spurk et al. 2020). Equation 1 describes the basic specification for a model with $k$ latent profiles and indicators $x_1 \ldots x_4$, where $P(x_1, x_2, x_3, x_4)$ is the overall probability of observing a particular response pattern for the indicators, $P(C_i)$ is the probability of an individual observation belonging in latent profile $C_i$, and $f(x_1, x_2, x_3, x_4 | C_i)$ is the multivariate probability density function for the indicators given latent profile $C_i$.

$$P(x_1, x_2, x_3, x_4) = \sum_{i=1}^{k} P(C_i) \cdot f(x_1, x_2, x_3, x_4 | C_i) \quad \text{(Equation 1)}$$

*Hypothesis Testing: Identifying Profile Structure*

Researchers recommend seeking a parsimonious and meaningful (or interpretable) model solution by specifying models that increase in terms of the number of profiles and comparing the respective model statistics and results (Spurk et al. 2020). We specified one- to five-profile models and compared the respective model statistics, as well as the interpretability of resulting profiles. Although the intraclass correlation coefficients for each of the four indicators are very low (< 0.07), for additional robustness, we computed standard errors that are adjusted for the country-strata. The full model statistics for one- to five-profile models are included in Web Appendix D. We sought the optimum tradeoff between model fit and complexity by comparing the Bayesian Information Criterion (BIC) values against model complexity (i.e., number of profiles). The BIC plot in Web Appendix E shows that the optimum model is a four-profile model, which also yielded an excellent degree of profile separation (entropy = 0.97). We confirmed the reliability of the four-profile solution by repeating the analysis with 100 random starts (Spurk et al. 2020). Additionally, we performed a Vuong-Lo-Mendell-Rubin robust likelihood ratio test, which showed that a four-profile model is significantly different (as well as better fitting) compared to a three-profile model ($\Delta\chi^2$/df = 37329.5/5; p=0.000). Finally, as visualized in Figure 2, the marginal means for the indicators in each profile also reflect the theorized attitude structure. Thus, H1 is supported.

**Figure 2. Attitude Profiles Derived from Archival Data (Study 1)**

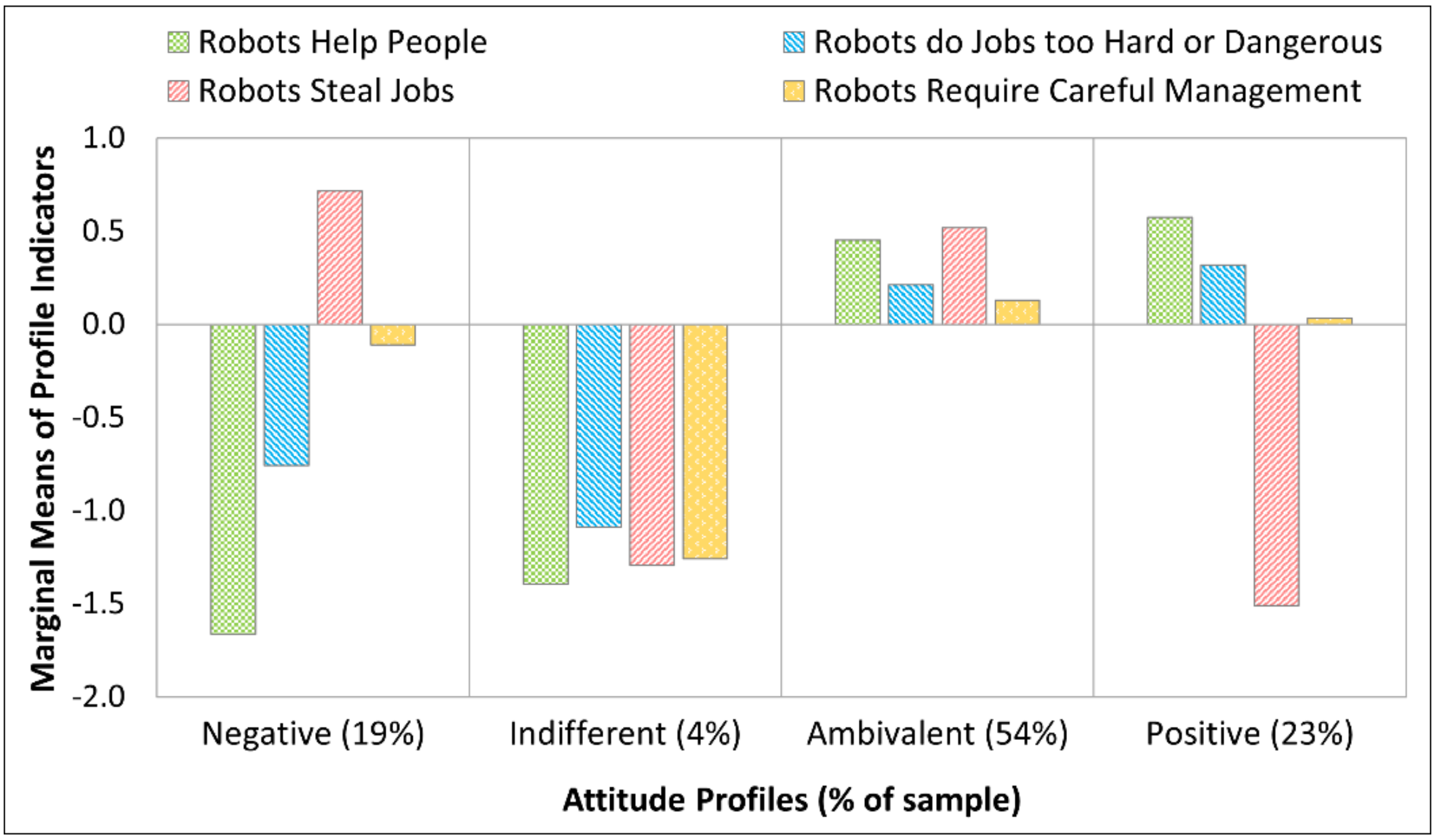

*Examining Profile Stability*

Subsequently, we specified a multigroup latent profile model with the latent profile means and all other parameters free to vary between the three groups representing the three waves of data collection (2012, 2014, and 2017). This (unconstrained) model was compared to a model with the profile means constrained across the three waves of data collection (constrained model), which revealed that the two models are equivalent ($\Delta\chi^2$/df = -3283.53/46; p = 1.00). Moreover, the constrained model (BIC: 806469.6) offers a better fit than the unconstrained model (BIC: 821195.6). Thus, the profile structure is stable across the three waves of data collection, supporting H2. Additionally, we found the same pattern of attitude profile means across two country-level cultural clusters (see Web Appendix F).

*Influence of Demographics and Technology Use on Profile Membership*

We sought to understand whether observed individual characteristics can help predict attitude profile membership. To this end, we reran the original latent profile model with demographic (*age*, *age at full-time education ceased, sex, occupation*) and technology-related factors (*internet usage*, *prior experience with robots*) incorporated as predictors of profile membership. The specifications for the model incorporating profile membership predictors $z_1 \ldots z_n$ can be described as in Equations 2 and 3, where $C_k$ is the base group for pairwise comparisons, $P(C_i|z_1, z_2, \ldots, z_n)$ represents the probability of belonging to latent profile $C_i$ as influenced by the predictors $z_1 \ldots z_n$, and the overall probability of observing a particular response pattern for the indicators $x_1 \ldots x_4$ is denoted by $P(x_1, x_2, x_3, x_4)$.

$$log\left(\frac{P(C_i|z_1, z_2, \ldots, z_n)}{P(C_k|z_1, z_2, \ldots, z_n)}\right) = \alpha_{1i} + \beta_{1i} \cdot z_1 + \beta_{2i} \cdot z_2 + \ldots + \beta_{ni} \cdot z_n \quad \text{(Equation 2)}$$

$$P(x_1, x_2, x_3, x_4) = \sum_{i=1}^{k} P(C_i|z_1, z_2, \ldots, z_n) \cdot f(x_1, x_2, x_3, x_4|C_i) \quad \text{(Equation 3)}$$

The details of the results are provided in Web Appendix G. A plot of effect sizes (relative risk ratios) for each predictor's influence on the likelihood ("risk") of observations belonging to each profile is provided in Web Appendix H. According to the results, females (vs. males) are *more* likely, while individuals with high internet use and prior experience with robots are *less* likely, to exhibit negative, indifferent, or ambivalent attitudes toward robots than positive attitudes. However, the demographic and technology-related factors are less helpful in differentiating between the four attitude profiles, since the same pattern of effects can be observed for all three other profiles compared to the positive profile.

*(In)Stability of Individual-level Attitudes and the Influence of Prior Experience*

Although the preceding analyses illustrate the stability of population-level attitudes (attitude profiles), the stability of individual-level attitudes is yet to be formally tested. To test this, we deployed a Double Machine Learning treatment effects model, which is a method of estimating the average causal effect of an uncontrolled and non-randomized 'treatment' while accounting for selection bias and overfitting (Chernozhukov et al. 2018). Accordingly, our model specification controlled for the effect of respondents' *age*, *age at full-time education ceased, sex, occupation,* and *internet usage* on both the treatment (i.e., prior experience with robots) and outcome (i.e., general attitude toward robots, measured on a 5-point scale; *M* = 3.44, SD = 1.15) variables. The analysis was repeated for each wave of data, and cluster-robust standard errors at the country level were computed for added robustness. Equations 4-6 describe the basic model, where $z_1 \ldots z_n$ are controls (demographics and internet use) selected using the LASSO method, and *ATE* is the average causal effect of individuals' prior experience on their attitude toward robots:

$$GeneralAttitude_i = \acute{\beta}_0 + \acute{\beta}_{1i} \cdot z_1 + \acute{\beta}_{2i} \cdot z_2 + \ldots + \acute{\beta}_{ni} \cdot z_n + \acute{\varepsilon}_i \quad \text{(Equation 4)}$$

$$P(PriorExperience_i = 1 | z_1, z_2, \ldots, z_n) = \frac{1}{1 + e^{-(\alpha_0 + \alpha_{1i} \cdot z_1 + \alpha_{2i} \cdot z_2 + \ldots + \alpha_{ni} \cdot z_n)}} \quad \text{(Equation 5)}$$

$$ATE = E[GeneralAttitude_i | PriorExperience_i = 1] - E[GeneralAttitude_i | PriorExperience_i = 0] \quad \text{(Equation 6)}$$

The results show that the *ATE* is positive and significant while controlling for the effects of demographics and internet use on both the treatment and outcome variables. Thus, individuals' prior experience with robots has a positive effect on their subsequent attitude toward robots, as per the results from 2012 (B = 0.47, p = 0.000), 2014 (B = 0.43, p = 0.000), and 2017 (B = 0.39, p = 0.000) survey waves.

**Study 2: Design and Measures**

To replicate the attitude structure in a new sample and to investigate the theory-based antecedents of attitude profiles, we recruited 950 participants from the USA via Prolific (mode age-group: 26-33 years; 51.9% female). The sample size comfortably exceeds expert recommendations for comparable model specifications (Spurk et al. 2020). We measured the original four benefit and risk perceptions (S1) as indicators of attitude profiles. In addition, we measured the psychological antecedents of profile membership using existing pre-validated scales as follows: 1) need for interaction with service employees (Dabholkar 1996); 2) basic psychological needs (autonomy, competence, relatedness) in the use of technology (Bergdahl et al. 2023). Scale items and reliability scores are provided in Web Appendix I.

**Study 2: Analysis and Results**

As in Study 1, a four-profile model emerged as the optimum solution based on a range of criteria. The BIC plot (reflecting model fit) for all models and the accompanying statistics are provided in Web Appendices J and K. The resulting marginal means for each attitude profile (see Web Appendix L) reveal a pattern consistent with Study 1. Subsequently, we incorporated the theory-based predictors into the latent profile model (as described in Equations 2 and 3 earlier). Table 2 summarizes the results. The need for autonomy increases likelihood of the negative (vs. positive) profile (supporting H3a), but there is no evidence for the same effect regarding the indifferent or ambivalent profiles (no support for H3b and H3c). The need for competence does not exert any significant influences; thus, H4a-H4c are not supported. In contrast, the need for relatedness increases the likelihood of negative, indifferent, and ambivalent profiles compared to the positive profile. Further, the negative (vs. positive) and ambivalent (vs. positive) profiles are more likely to occur due to greater need for interaction with service staff. So, H6a and H6b are supported, but H6c is not.

**Table 2. Antecedent Effects on Attitude Profile Membership (Study 2)**

| **Attitude Profile Comparison** | **Need for Autonomy** B [95% CI] | **Need for Competence** B [95% CI] | **Need for Relatedness** B [95% CI] | **Need for Interaction with Staff** B [95% CI] |
|---|---|---|---|---|
| Ambivalent vs. Positive | 0.10 [-0.29, 0.49] | -0.06 [-0.41, 0.29] | 0.51* [0.07, 0.95] | 0.90* [0.50, 1.30] |
| Indifferent vs. Positive | 0.36 [-0.00, 0.71] | -0.04 [-0.38, 0.30] | 0.48* [0.03, 0.92] | 0.18 [-0.29, 0.49] |
| Negative vs. Positive | 0.60* [0.24, 0.97] | -0.01 -0.35, 0.34] | 0.86* [0.41, 1.30] | 0.84* [0.39, 1.29] |
| Indifferent vs. Negative | -0.25 [-0.57, 0.08] | -0.03 [-0.35, 0.30] | -0.38* [-0.72, -0.04] | -0.67* [-0.97, -0.36] |
| Ambivalent vs. Negative | -0.50* [-0.81, -0.20] | -0.05 [-0.34, 0.25] | -0.35* [-0.64, -0.05] | 0.05 [-0.30, 0.41] |
| Indifferent vs. Ambivalent | 0.26 [-0.05, 0.57] | 0.02 [-0.27, 0.31] | -0.04 [-0.35, 0.28] | -0.72* [-1.10, -0.34] |

**Discussion**

Considering the uniqueness of service robots as an innovation, juxtaposed with the general negativity surrounding AI and robots in popular culture (Gnambs and Appel 2019), the existence and stability of a substantially positive attitude profile is indeed noteworthy. The latter finding supports the rapidly expanding implementation of robots in service frontlines. In addition, while nascent research has observed ambivalent attitudes (e.g., Dang and Liu 2021), indifferent attitudes towards service robots have remained unacknowledged until now. Furthermore, although prior research has established negative attitudes towards service robots (Luo et al. 2019), our findings show that only one subsection of the population (< 20%) represents a generally unfavorable disposition. While this proportion of negative attitudes is not negligible, clearly the largest representation of attitudes at population level are ambivalent or indifferent profiles, which supports and extends the findings of Stapels and Eyssel (2021).

Study 2 reproduced the profile structure and added evidence as to its stability. In terms of psychological antecedents of attitude profiles, the needs for relatedness and interaction with service staff emerge as important predictors, but not the need for competence. For instance, the key difference between the indifferent and positive profiles is the need for relatedness, while the need for interaction with service staff is the antecedent that helps distinguish between the ambivalent and negative profiles. The importance of the needs for relatedness and interaction with service staff as attitude profile antecedents suggests that, even as robots are increasingly implemented in service frontlines, a substantial part of the population still values direct human contact. These findings align with prior research showing that robot-related positive socio-cognitive effects fail to materialize in the co-presence of human service staff (Yoganathan et al. 2021). The absence of differential predictive effects of the need for competence suggests that people are more used to dealing with AI and service robots owing to their rapid proliferation. This is also reflected in the limited influence of the need for autonomy, which only reveals significant effects where the negative profile is concerned (cf. Jörling et al. 2019; Odekerken-Schröder et al. 2020).

## STUDIES 3 AND 4: ATTITUDE OUTCOMES BASED ON REAL INTERACTIONS

### Study 3: Design and Measures

To develop this study, we collaborated with UneeQ, a company that elevates enterprise brand engagement and experience through digital humans – i.e., humanlike intelligent interfaces capable of real-time interactions. UneeQ digital humans are deployed by many leading brands worldwide (https://www.digitalhumans.com). We programmed a separate chatbot using Google Dialogflow to manage a service task, whereby a user would ask the chatbot to find hotel prices for a location they would like to visit. The programmed chatbot was then integrated into the UneeQ interface, which allows users to interact with the digital human to complete the service task for the study. The digital human uses our programming of the chatbot as the 'rules of engagement' to interact with users. Further details of the virtual robot and the chat flow are provided in Web Appendix M. We recruited 301 participants from Australia through Prolific (mode age-group: 26-33 years; 54.5% female). We measured the original four attitudinal indicators (S1) and participants' pre-existing level of experience with service robots, as well as demographics. Participants were then instructed that they were planning to visit London (UK), and that the digital human ("Alex") would help them search for a hotel (see chat flow in Web Appendix M). After the interaction, participants recorded their discomfort and anxiety about interacting with Alex in the future on a 7-point semantic differential scale adapted from Yoganathan et al. (2021).

### Study 3: Analysis and Results

We specified a latent profile model as in previous studies (Equation 1), but integrated a path model with participants' discomfort and anxiety regarding future service robot interaction as dependent variables while controlling for prior experience. The integrated model estimated the response probabilities in each latent profile for the indicators and the dependent variables *discomfort* and *anxiety* (controlling for prior experience) using maximum likelihood estimation. The results (see Figure 3) reveal that post-interaction discomfort and anxiety conform to expectations – e.g., positive/negative profiles display lowest/highest levels of discomfort and anxiety, and the indifferent profile's means are statistically not different from zero on a standardized scale (i.e., no notable reaction to the robot). Corresponding to a relatively higher level of risk perception, the ambivalent (vs. indifferent) profile displays greater discomfort (t = 20.8; p = 0.00) and anxiety (t = 8.9; p = 0.00).

**Figure 3. Post-interaction Discomfort and Anxiety (Study 3)**

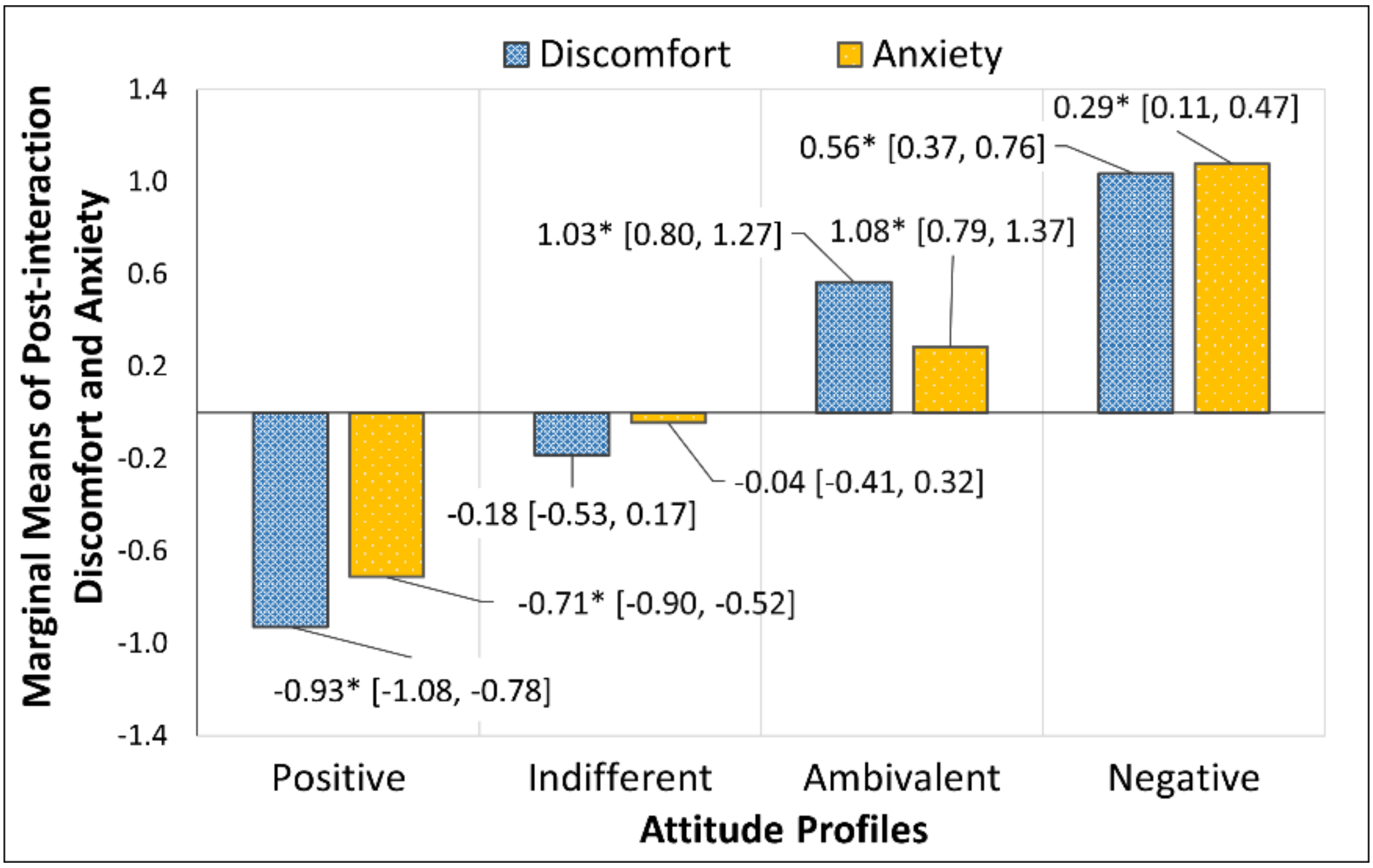

We conducted additional analyses to check whether the interaction with the digital human alters individual-level attitudes. We used the mean values of responses to the four attitudinal variables (profile indicators) to represent individuals' pre-interaction overall attitude toward robots. The responses for discomfort were reversed to obtain participants' post-interaction comfort level. We then conducted a Wilcoxon signed-rank test (nonparametric approach due to difference in scales), while controlling for participants' pre-existing experience with service robots. Results revealed a statistically significant increase in the post-interaction comfort (vs. pre-interaction attitude) for three levels of pre-existing experience with service robots: low (i.e., Quartile 1: $z = 9.78$; $p = 0.000$), medium (i.e., Quartile 2: $z = 8.46$; $p = 0.000$), and high (i.e., Quartile 3: $z = 7.72$; $p = 0.000$) experience.

**Study 4: Data and Measures**

We extracted English language online reviews of 20 hotels worldwide that are known for using service robots (see Web Appendix N). A total of 4,696 reviews containing the word 'robot' were extracted from three different platforms: Google, TripAdvisor, and Booking.com. Advances in computational linguistics have enabled researchers to identify specific psychological constructs from textual data (Berger et al. 2020). By applying the Certainty and Evaluative Lexicon methods (Rocklage et al. 2023; Rocklage, Rucker, and Nordgren 2018), we computed four dimensions of customer attitudes as reflected in textual online reviews: average *certainty*, *emotionality*, *extremity*, and *valence*. We also extracted the customer rating of the robot hotel associated with each review. Finally, the type of robot used in each hotel was identified from a combined examination of specific reviews mentioning the robots by name, the hotel website, and media reports pertaining to the hotel's use of robots.

**Study 4: Analysis and Results**

We specified a finite mixture model with *certainty*, *emotionality*, *extremity,* and *valence,* along with the *customer rating* for the hotel, as dependent variables. We controlled for the potential influence of the type of robot and platform (Google, TripAdvisor,

Booking.com). The overall probability of responses for dependent variables $y_1 \ldots y_n$ with $k$ latent profiles while controlling for robot and platform type is described as follows:

$$P(y_1, y_2 \ldots, y_n | Robot, Platform) = \sum_{i=1}^{k} P(C_i) \cdot f(y_1, y_2 \ldots, y_n | C_i, Robot, Platform) \quad \text{(Equation 7)}$$

Two types of additional robustness specifications were made to the model: 1) robust standard errors were computed to account for the nesting of dependent variables within hotels, and 2) the stability of the solution was tested by repeating the analysis with 100 random draws and 20 iterations per draw (i.e., to achieve global vs. local maxima). As in previous studies, the four-profile model revealed optimum model fit. The pattern of marginal means (adjusted for robot and platform type, and based on cluster-robust standard errors) are visualized in Figure 4. The ambivalent profile is characterized by the lowest level of certainty expressed in reviews, while the indifferent profile is characterized by relatively low and negative valence and emotionality. The negative and positive profiles also reveal an expected pattern of responses. As the behavioral outcome, we are interested in the overall rating accorded to the hotel based on attitude profile. The indifferent profile is again more positive (as in Study 3) compared to the ambivalent profile in terms of the overall ratings, while negative and positive profiles expectedly reveal the lowest and highest ratings.

**Figure 4. Attitude Profiles Derived from Online Reviews (Study 4)**

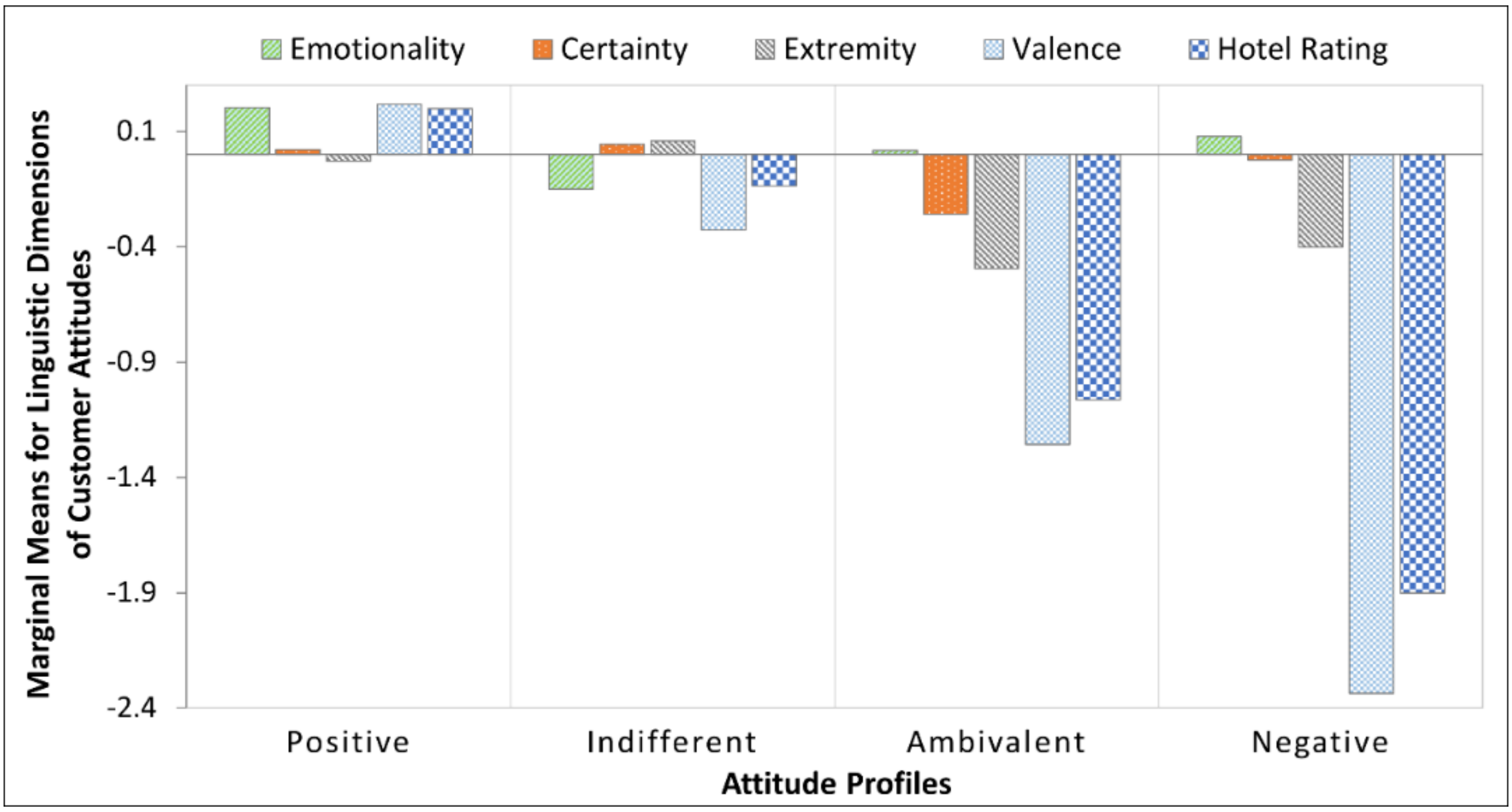

To derive further insights, we combined natural language processing with qualitative analysis of the review corpus. As per Table 3, we identified five themes: service quality, robot performance, room size and layout, facilities, and price. We then evaluated the general sentiment associated with these themes to identify whether customers within each attitude profile were satisfied/unsatisfied with each of the five aspects. This analysis helped understand the (dis)satisfaction with service robots in the context of various other salient factors (e.g., general service quality). The results are consistent with the characteristics of the attitude profiles identified in preceding studies, and further qualitative analysis of the reviews allowed us to identify specific managerial action points for each attitude profile (discussed next).

**Table 3. Key Issues by Attitude Profile and Recommendations for Hospitality Service Improvement (Study 4)**

| Attitude Profiles & Illustrative Quotes | Comparison of Key Issues* | Suggestions for Improvement (hospitality context) |
|---|---|---|
| **Negative** | | |
| *"I didn't like the robot service at all. I prefer talking to real people."*<br><br>*"The room was tiny and dark. Breakfast was disappointing, and the staff seemed inexperienced. The robots were more of a gimmick than actually useful."*<br><br>*"The robot delivery was annoying. The hotel should focus more on improving human service."* | • Service quality: 27.3% (5.5% positive, 21.8% negative).<br>• Robot performance: 17.9% (3.6% positive, 14.3% negative).<br>• Room size and layout: 12.4% (3.1% positive, 9.3% negative).<br>• Facilities: 7.8% (2.0% positive, 5.8% negative).<br>• Price: 26.8% (5.4% positive, 21.4% negative). | ➢ Increase staffing during peak check-in and check-out times (e.g., "roaming concierge"), and offer personalized welcome notes from human staff.<br>➢ Create guided tours highlighting non-tech features of the hotel, such as its architecture/design or heritage/history.<br>➢ Develop partnerships with local attractions for exclusive guest experiences (e.g., city tours, or Michelin-star dining experiences).<br>➢ Introduce "Tech-Free" spaces with traditional amenities. |
| **Ambivalent** | | |
| *"The robot was cute but didn't always work as expected. It would be nice to have a human option available."*<br><br>*"The hotel had some cool features like the robots and Alexa, but the room layout was impractical. The staff was friendly, but the service was inconsistent. It's an interesting concept, but not sure if I'd stay again."*<br><br>*"I was unsure about the robot at first, but it worked well most of the time. Still, I preferred interacting with the staff."* | • Service quality: 24.7% (14.8% positive, 9.9% negative).<br>• Robot performance: 43.6% (26.2% positive, 17.4% negative).<br>• Room size and layout: 42.3% (15.2% positive, 27.1% negative).<br>• Facilities: 49.0% (19.6% positive, 29.4% negative).<br>• Price: 26.8% (13.4% positive, 13.4% negative). | ➢ Create a dedicated concierge position for technology-related queries, especially available in common areas, and offer in-person orientations for guests unfamiliar or uncertain about hotel technology.<br>➢ Enable guests to customize aspects of their rooms, such as in-room amenities and ambience.<br>➢ Manage customer expectations by providing detailed pre-arrival information about the hotel's features and specific details about the room (e.g., virtual tours).<br>➢ Gather detailed feedback by implementing a comprehensive post-stay survey or informal interviews to understand mixed reactions. |

| Attitude Profiles & Illustrative Quotes | Comparison of Key Issues* | **Suggestions for Improvement** (hospitality context) |
|---|---|---|
| **Indifferent** | | |
| *"The hotel was fine, nothing special. The robot delivery was a neat gimmick but didn't really add much to the experience."*<br><br>*"Nice hotel with good basic amenities, but the robot was just a gimmick for us."*<br><br>*"The room was clean and comfortable, but the robot delivery service didn't add much to our experience."* | • Service quality: 22.1% (13.3% positive, 8.8% negative).<br>• Robot performance: 17.9% (10.7% positive, 7.2% negative).<br>• Room size and layout: 28.7% (10.5% positive, 18.2% negative).<br>• Facilities: 31.4% (12.6% positive, 18.8% negative).<br>• Price: 25.0% (10.0% positive, 15.0% negative). | ➢ Offers could focus on creating memorable experiences that take advantage of hotel's technology-focus – e.g., technology-enhanced experiences of the hotel, such as via virtual reality or metaverse, and using virtual robots.<br>➢ Train staff to explain benefits before recommending or redirecting to robot services in order to build interest and clarify added value.<br>➢ Improve communication of hotel amenities in general – e.g., create an interactive guide highlighting all unique features and traditional service excellence. |
| **Positive** | | |
| *"Loved everything about this hotel! The robots Leo and Cleo were so cool, and the staff was incredibly friendly. The room design was unique and comfortable. We'll definitely be back!"*<br><br>*"The robot delivery service was amazing! It made our stay so much fun."*<br><br>*"We loved the robots! They were efficient and added a unique touch to our stay."* | • Service quality: 25.9% (19.4% positive, 6.5% negative).<br>• Robot performance: 20.6% (16.5% positive, 4.1% negative).<br>• Room size and layout: 16.6% (9.1% positive, 7.5% negative).<br>• Facilities: 11.8% (7.1% positive, 4.7% negative).<br>• Price: 21.4% (15.0% positive, 6.4% negative). | ➢ Implement a monthly "new feature" rollout for robot services, including robot-guided tour options, and/or interactive games or challenges featuring robots.<br>➢ Develop an "Ambassador for Robots" program for frequent guests, which could also involve social media competitions for the most engaging user-generated content with robots to increase publicity.<br>➢ Create a mobile app for tracking and customizing customer interactions with hotel robots, which could also be useful for sustaining the engagement beyond the visit. |

*Note*. *based on mentions within attitude profile as a % of all mentions.

**Discussion**

Based on actual interactions and online reviews of robot hotels from different countries, we revealed that the four attitude profiles manifest in the 'real world' and predict customer evaluations and satisfaction ratings. The comparative effects of ambivalent and indifferent profiles on post-interaction discomfort and anxiety and overall hotel ratings, illustrate that the indifferent (vs. ambivalent) profile's behavior is relatively positive, while ambivalent is more comparable to the negative profile. This indicates that risk perceptions (*ambivalent* displays higher perceived risk as per S1 and S2) play a more influential role, compared to benefit perceptions, in shaping customer evaluations. Considering that the ambivalent (vs. indifferent) profile exhibits greater need for interaction with service staff (S2), our findings also advance previous research showing a strong relationship between the need for interaction with service staff and robot-related perceived risk (Yoganathan et al. 2021). Although post-interaction discomfort and anxiety appear to differ only in magnitude between the indifferent and positive profiles, the indifferent profile's means are non-significant, which reflects this profile's apathetic stance regarding service robots. In this regard, our findings complement those of other researchers (Stapels and Eyssel 2021), but also advance the discussion further in terms of attitude structure (population vs. individual levels) and approaches to measurement (different measures, actual interactions, and textual data).

As explained in Table 3, our analyses of online reviews revealed further insights into how different attitude profiles evaluate various aspects of staying in a robot hotel. For example, to mitigate unfavorable evaluations of the negative profile, human staffing levels can be increased during peak check-in/check-out times, and the non-robot or non-technological selling points of the hotel can be emphasized (e.g., exclusive tours or dining experiences, exploration of the hotel's heritage/history or cultural aspects). For the ambivalent profile, hotels may introduce a 'tech concierge' and in-person orientations to alleviate uncertainty about the robot services. Further, managing expectations through pre-arrival information and post-stay feedback mechanisms would help understand mixed attitudes. For the indifferent profile, focusing on creating memorable experiences using technology (e.g., virtual reality or metaverse, and virtual robots), as well as better communicating the non-robot attractions and service excellence of the hotel is likely to be effective. Training staff to explain benefits before recommending or redirecting to robot services is also advisable to build interest among the indifferent. Finally, for the positive profile, it is important to maintain their excitement about robot services. So, hotels could roll out new features for robot services, including robot-guided tours and/or interactive games. Perhaps a loyalty program to reward frequent customers, which also leverages their excitement to increase publicity (e.g., via social media) for hotel robots would yield multiple benefits for hotels as well as 'positive' customers.

## STUDY 5: EFFECTS OF ANTHROPOMORPHIC ROBOT DESIGN

### Data and Measures

We utilized the full dataset from Spatola and Wykowska's (2021) research, which provided several advantages. First, the dataset contains 1,141 participants' evaluations of five types of robots – thus providing a large enough sample, as well as evaluations of multiple types of robots. Second, the data is part of an open science project[1]. Finally, replicating our attitude structure in diverse samples/contexts, and using different measures, would increase its robustness and generalizability (cf. Dang and Liu 2021; Stapels and Eyssel 2021). Spatola and Wykowska (2021) selected five specific robots to represent increasing degrees of humanlike appearance based on the measurements contained in the ABOT database (Phillips et al. 2018). After viewing the pictures of each robot in random order, participants rated the robots' agency (ability to think and act), sociability (friendliness or warmth), and uncanniness (causing creepiness or uneasiness) on a 7-point scale. They also completed a six-item version of the Negative Attitudes towards Robots Scale (NARS).

[1] European Union Horizon 2020 program, ERC grant, G.A. number: ERC-2016-StG-715058.

**Analysis and Results**

Following the same approach as in Study 3, we specified a path model integrated into a latent profile model using the NARS items as indicators. The path model included perceived agency, sociability, and uncanniness as dependents and the type of robot as predictor. A four-profile model was chosen following the same procedure as in our previous studies (see Web Appendix O for profile plot). The perceived sociability and uncanniness based on robot type are visualized in Figure 5 for each attitude profile. In terms of uncanniness, the differences between profiles become more pronounced at the high end of the anthropomorphic spectrum, with the highly humanlike robot ("Nadine") eliciting the highest (lowest) level of uncanniness in the negative (positive) profile. The indifferent profile displays a lower level of uncanniness than the ambivalent profile for Nadine. The lowest level of uncanniness for all profiles is associated with the humanoid robot "Nimbro" (fourth from the left in Figure 5), which is not as humanlike in design as Nadine. Notably, the effects for perceived sociability by robot design are almost a mirror image of those observed for uncanniness, with Nimbro registering the highest level of sociability for all profiles. Perceived agency increases with the anthropomorphic design of the robot in all attitude profiles, with few pronounced differences between the profiles in the extent to which this increase occurs (Web Appendix P). Irrespective of the robot type, we observe similarities between ambivalent and positive profiles in average perceived sociability (i.e., the average of perceived sociability values across five service robots; see Web Appendix Q). In contrast, the ambivalent and negative profiles are similar in terms of average uncanniness. Further, the indifferent (vs. negative) profile is significantly *higher* in sociability and *lower* in uncanniness. Taken together, these observations provide a further explanation as to why the indifferent (vs. ambivalent) profile's outcomes in our Studies 3 and 4 are more positive.

**Figure 5. Robot Type and Anthropomorphic Effects by Attitude Profile (Study 5)**

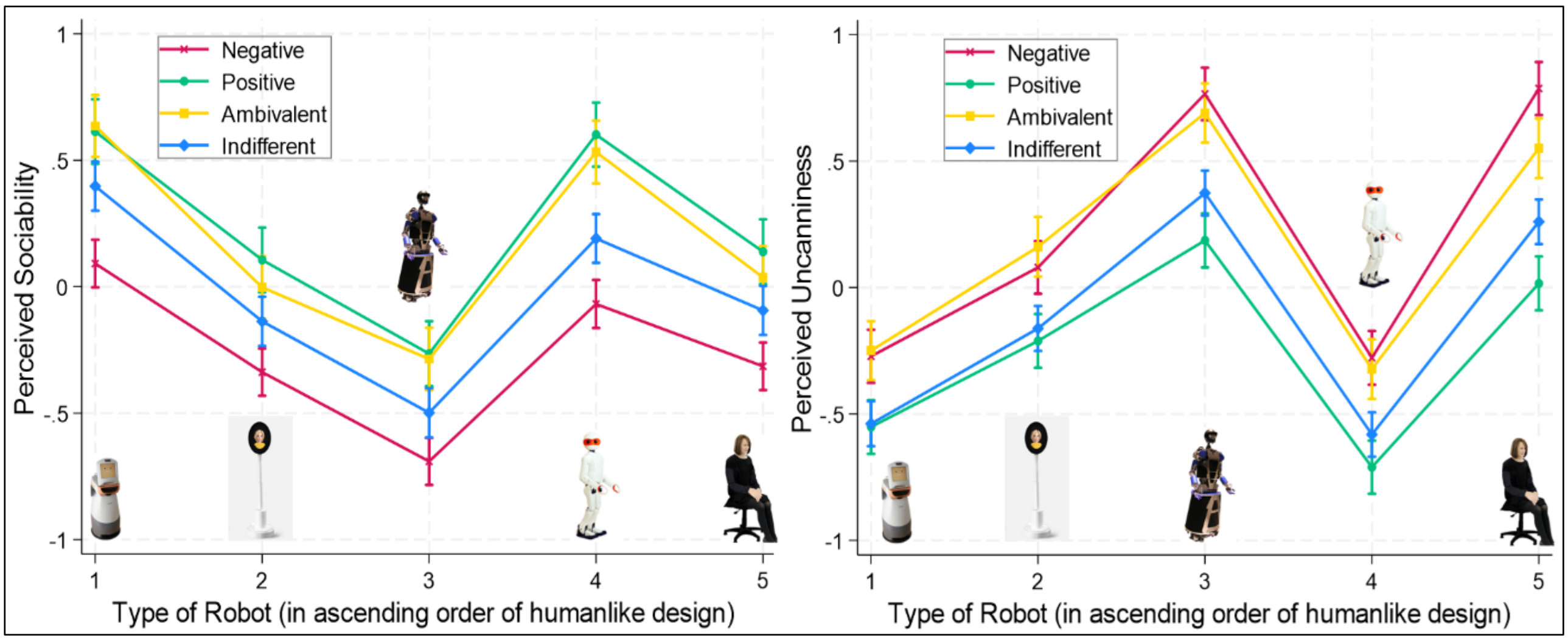

## Discussion

Findings from Study 5 extend conclusions reached by Spatola and Wykowska (2021), illustrating how the population-level attitude structure gives rise to different socio-cognitive evaluations of the same service robot. Although the indifferent (vs. ambivalent) profile's average tendency for perceiving sociability in a service robot is significantly lower, the indifferent (vs. ambivalent) profile also has a significantly lower tendency for perceiving uncanniness. The indifferent and positive profiles display similar levels of average uncanniness, while the ambivalent profile is similar to the negative profile. This implies that the different baseline tendencies for anthropomorphic evaluations (irrespective of robot type) underlie the indifferent (vs. ambivalent) profile's relatively positive behavioral outcomes. In terms of robot design, extremely humanlike robots such as Nadine elicit the worst evaluations, whereas the relatively less humanoid Nimbro elicits the best evaluations across all profiles, which adds nuance to prior research showing positive evaluations of humanoid robots (Christou, Simillidou, and Stylianou 2020; Yam et al. 2021).

# GENERAL DISCUSSION

## Overview of Key Findings

Building on theory and prior research on attitude structure and stability, we argued for the importance of understanding attitudes toward service robots at the population level, and subsequently provided empirical evidence and in-depth explanations for this through five diverse studies. A graphical summary of the research gaps addressed, and key findings from each study are provided in Figure 6.

**Figure 6. Graphical summary of research gaps addressed and key findings**

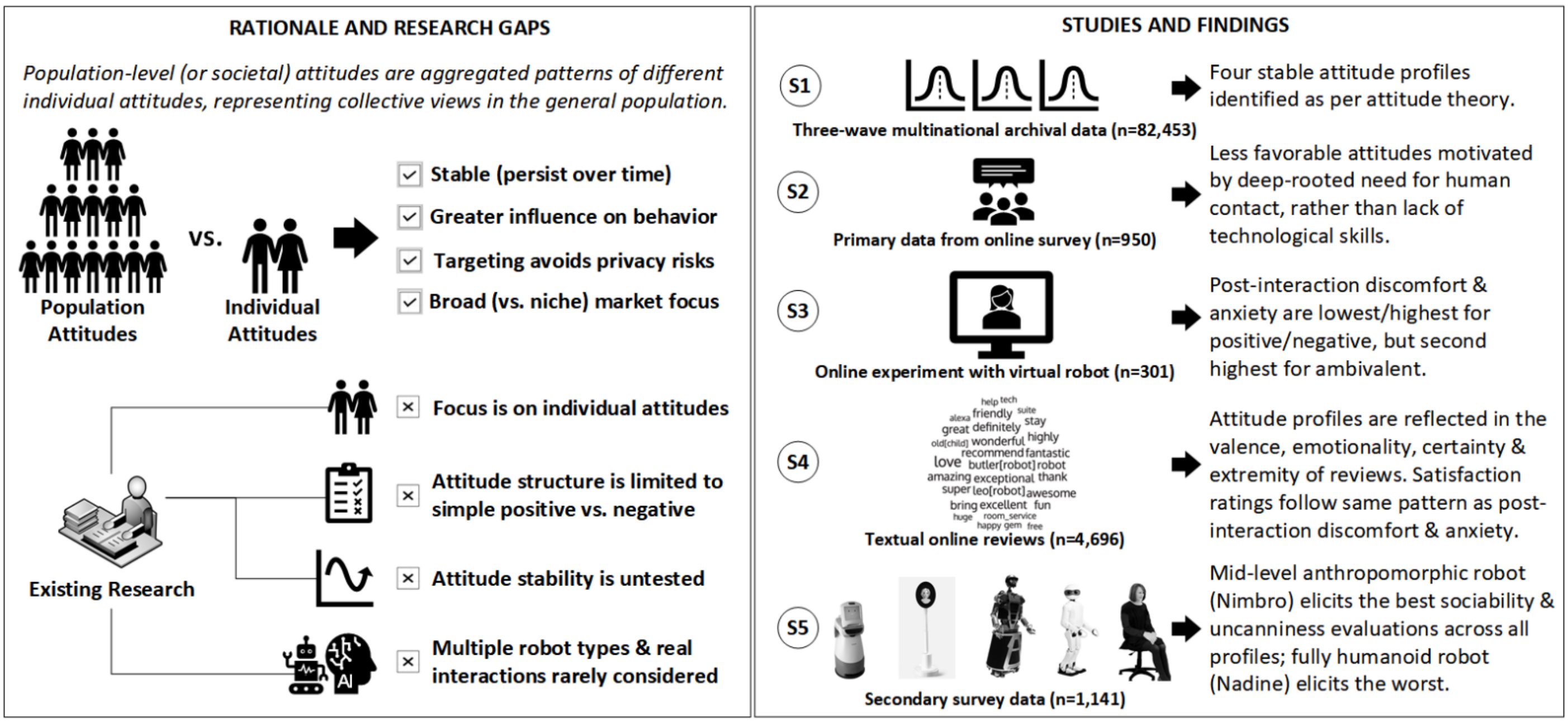

Based on the predictions of attitude theory, Study 1 uncovered four distinct and stable attitude profiles by utilizing multinational data collected in three waves: *positive* (low risk and high benefit perceptions), *negative* (high risk and low benefit perceptions), *indifferent* (low risk and low benefit perceptions), and *ambivalent* (high risk and high benefit perceptions). The stability of these profiles was established across Studies 1 to 5, as well as across cultural clusters of countries (S1), in that the same attitude structure and profile characteristics were observed repeatedly. In addition, we also showed that individuals' attitudes change based on personal experience with service robots (S1 and S3).

Study 2 revealed that the need for relatedness and autonomy in technology use, and the need for interaction with service staff, influence the membership of the attitude profiles. The likelihood of negative (vs. positive), indifferent (vs. positive), and ambivalent (vs. positive) profiles increases with the need for relatedness. Similarly, the likelihood of negative (vs. positive) and ambivalent (vs. positive) profiles increases with the need for interaction with service staff. Notably, the need for autonomy and competence in the use of technology have little or no influence on attitude profile membership.

In Studies 3 and 4, based on actual interactions with service robots, we showed that the four attitude profiles predict differences in post-interaction discomfort and anxiety (S3), and satisfaction ratings and service evaluations of robot hotels (S4). These results established the real-world relevance of the attitude profiles, but also added nuance to the profile characteristics. The indifferent profile's influence on customer behavioral patterns is closer to that of the positive profile, while the ambivalent profile was more comparable to the negative profile in this regard.

Study 5 shed light on how the different profiles evaluated robots' humanlike designs, with the highly humanlike robot eliciting the worst evaluations across all profiles and the robot with the relatively low humanlikeness eliciting the best evaluations. Study 5 also provided a possible underlying explanation for the indifferent profile's alignment with the positive profile, and the ambivalent profile's alignment with the negative profile. While the indifferent and positive profiles reveal comparable tendencies on average for perceiving uncanniness, the ambivalent profile is more akin to the negative profile.

Thus, our large-scale investigation on population-level attitudes reveals a stable structure of four profiles, which predict significant differences in customer evaluations of robot-delivered services, as well as the robots that deliver those services. From a psychological perspective, even as people become more used to interacting with service robots, our findings suggest that the less positive attitudes at population-level emerge out of a deep-seated motivation for seeking human contact. The theoretical and managerial implications of our findings are summarized in Figure 7.

## Figure 7. Theoretical and Managerial Implications

### THEORETICAL IMPLICATIONS

**Profile Structure and Stability**

- While individual attitudes change based on experience of interacting with robots, the attitude structure at population level remains stable.
- Diverse samples and different measures consistently reveal a structure of four attitude profiles: positive (favorable predisposition), indifferent (low involvement), ambivalent (high uncertainty), and negative (unfavorable predisposition).
- This stable structure provides an *a priori* theoretical basis for classifying population-level heterogeneity in attitudes to service robots – i.e., future research should account for the underlying structure of four attitude profiles in the empirical examination of customer behavior or evaluations.

**Psychological Antecedents**

- The need to connect or interact with other people (including, but not limited to, service employees) is a fundamental driver of less positive attitudes (i.e., indifferent, ambivalent, or negative) towards robots.
- The need for competence does not differentiate between attitude profiles. The need for autonomy plays a limited role (high in the negative profile compared to ambivalent and positive profiles). These observations may reflect increasing customer familiarity with AI applications.

**Customer Outcomes**

- Attitude profiles predict consistent differences in post-interaction discomfort and anxiety, as well as satisfaction ratings and assessment of service features (e.g., service quality, robot performance).
- The indifferent profile exhibits more positive reactions than the ambivalent profile, indicative of risk perceptions playing a more influential role than benefit perceptions in shaping customer evaluations.
- Attitude profiles reveal different thresholds for perceiving sociability and uncanniness in robots, which helps explain profile differences in discomfort, anxiety, and satisfaction.
- The indifferent and positive profiles display comparably similar levels of perceived uncanniness (irrespective of robot type), whereas the ambivalent profile is similar to the negative profile.
- Extremely humanlike robots (e.g., Nadine) elicit the worst sociability, agency, and uncanniness perceptions across all profiles.

### MANAGERIAL INSIGHTS & RECOMMENDATIONS

**Positive**

- Enhance robot services to keep customers engaged. This may involve technological upgrades (such as pairing physical robots with large language models), or new service features using the robot (e.g., a personal robot butler).
- Implement loyalty programs to reward repeat visitors with special benefits, such as exclusive interactions with high-end robots.
- Leverage positive customer experiences for organic marketing – for example, an 'Ambassador for Robots' program offering special benefits in return for sharing interactions and experiences with robots on social media.

**Negative**

- Emphasize human interaction by ensuring that human staff are available and visible to address needs, and by integrating social events into the core service offering. Support the need for autonomy by enabling customers to choose between robot vs. human assistance.
- Highlight non-robot features by drawing attention to other unique and value-adding aspects of a service (e.g., social aspects, tradition, ambience/experience).
- Staff training should ensure that, while maintaining the quality of hybrid services to create seamless experiences, customers do not feel compelled to use robot services.

**ATTITUDE PROFILES**

- Targeting population (vs. individual) level attitudes is beneficial for firms due to population-level attitudes' stability and stronger influence on customer behavior.
- This approach can also help firms overcome psychological barriers, such as privacy concern, owing to its limited reliance on personal customer data.
- Robots with median humanlikeness (e.g., Nimbro) may appeal to the broadest range of customer attitudes, based on customer perceived sociability, agency, and uncanniness.

**Indifferent**

- Service design should focus on creating memorable service experiences, of which the robots are a constituent – e.g., a farm-to-table dining experience where robots provide an interactive guide to the meal (explaining its provenance, ingredients etc.)
- Avoid overemphasizing robot services in marketing campaigns, and highlight other attractions/benefits that enhance the overall experience.
- Where robot services are referred to in marketing materials, explain how they improve overall service quality as opposed to framing them as the centerpiece.

**Ambivalent**

- Balance robot services with conventional value-adding elements (e.g., hotel room comfort), so that technological features enhance rather than complicate customer experiences.
- Ensure the availability of human staff, and include opportunities for socializing to enhance the overall service offering.
- Increase reliability of robot performance through a tracking system to detect and pre-empt performance issues and malfunctions. When updates/upgrades are made to robot systems, clearly communicate changes in functionality to customers to avoid exacerbating any uncertainty about robot-delivered service features.

**Theoretical Implications**

Despite attitude theory suggesting a multidimensional structure of attitudes toward service robots at population level (Pidgeon et al. 2005), this has been largely overlooked in prior research. Ignoring attitudinal heterogeneity produces biased or inconsistent findings and impedes effective interventions against adoption barriers (Becker et al. 2013; Bryan et al. 2021), especially where advanced AI services are concerned (see: Osburg et al. 2022). Yet, until now, empirical evidence has been lacking on the specific nature of the unobserved heterogeneity in attitudes toward service robots. Our research addresses this, and further explains the antecedents and consequences of population-level attitudes.

The four profiles reveal distinct characteristics based on quantitative (e.g., certainty and emotionality scores in textual reviews) and qualitative (e.g., comments about service robots) measures. It is, however, important to consider the attitude structure as a whole, as well as each attitude profile individually. As summarized in Figure 7, examining all profiles together helps differentiate between their behavioral outcomes. It also helps understand the relative significance of each attitude profile in terms of size – e.g., the combined indifferent and ambivalent profiles represent a larger proportion of the population than the combined positive and negative profiles.

From a broader perspective, our results illustrate that the need for human contact through connecting or interacting with other people (including, but not limited to, service employees) is a fundamental driver of less positive attitudes (indifferent, ambivalent, or negative) towards robots, as opposed to individual need for competence or autonomy in the use of technology (cf. Bergdahl et al. 2023). This also presents an interesting situation in light of the rise of empathetic or “feeling” service robots and their positive impact on wellbeing (Mende et al. 2024). Previous research offers some indication that the so called ‘feeling robots' may be able to fulfill the innate desire for human connection displayed by negative, ambivalent, and indifferent profiles – for example, Gelbrich et al. (2021) found that the emotional support received from an AI assistant increased customer perceived warmth.

By elucidating whether the attitude structure is multidimensional, and how many stable attitude profiles exist, we provide the methodological groundwork for designing (as well as analyzing) future studies. For example, a multidimensional attitude structure cannot be sufficiently accounted for by using a single scale (Dang and Liu 2021; Stapels and Eyssel 2021), and researchers require *a priori* theoretical guidance for determining the number of (unobserved) attitude profiles in a statistical model (Spurk et al. 2020). Moreover, as opposed to the arbitrary categorization of attitudes based on *theoretical* constructs (e.g., low vs. high negativity to robots based on NARS), our approach enables the identification of naturally occurring attitudes in the population. As we have illustrated, a finite mixture modeling facilitates not only the incorporation of the underlying heterogeneity into a path model, but also the comparative assessment of different model specifications (Osburg et al. 2022). This helps further fine-tune statistical models to suit sample-specific characteristics. For example, a three-profile structure may fit better for samples that are skewed towards high experience with service robots or contain too few observations exhibiting a specific attitude. Thus, we develop a theoretical foundation for understanding heterogeneous attitudes toward service robots, enabling future researchers to derive sound hypotheses and avoid spurious findings .

**Managerial Implications**

As service robots become more mainstream, there is greater need for firms to acknowledge and cater to different population-level attitudes. It is also impractical for service firms to target individual preferences, since the variability in individual customer attitudes may well be exacerbated by the fast and dynamic nature of developments in service robot technology. Furthermore, highly individualized robot services may face difficulties due to privacy and ethical concerns, which are driving calls for greater legislative control (Puntoni et al. 2021). So, while customers are reluctant to interact with service robots due to concerns about privacy or data protection (Čaić et al. 2018), regulators may also be compelled to enforce protective measures that curtail firms' ability to leverage customer data (Puntoni et al. 2021). Instead, targeting population-level attitudes (which are more stable) can be more feasible and effective in overcoming psychological barriers (e.g., privacy concerns), as there is limited need for personal customer data. Firms can assess robot-related customer risk and benefit perceptions to infer and target their corresponding attitude profile at a given time.

For the negative profile, considering the relatively higher need for human contact/connection, firms should emphasize the availability of human staff and create socializing opportunities as part of the core service offering. For example, M Social Singapore has a restaurant with beautiful artwork, an open kitchen, and big communal tables designed to encourage conversation. Further, staff training should focus on maintaining the quality of hybrid (human-technology) services for a seamless experience, while simultaneously ensuring that customers in the negative profile do not feel forced to use robot services (in light of their relatively higher need for autonomy).

For the ambivalent profile, it is important to balance technology with conventional aspects of services that add value, while alleviating uncertainties *vis-à-vis* robot services. A tracking system to detect and prevent robot issues or malfunctions can help reduce uncertainty about robot services and increase their reliability. In parallel, the availability of human staff should be ensured, as it is likely to appeal to this profile's psychological needs. Social spaces could be used to enhance the overall service offering from a non-technology perspective. Yotel New York City, for example, has a rooftop terrace called *The Social Drink and Food* for cocktail parties and events with great city views.

For the indifferent profile, service design should focus on creating memorable service experiences that include robots – e.g., a farm-to-table dining experience where robots provide an interactive guide to the meal (about its provenance, ingredients etc.). However, firms should avoid overemphasizing robot services in marketing campaigns, and highlight other attractions/benefits (e.g., location, amenities) instead. Where robot services are referred to in marketing materials, an explanation of how these enhance overall service quality should be provided, as opposed to positioning them as the centerpiece.

For customers exhibiting positive attitudes, enhancements to robot services can help keep them engaged. Enhancement could be technological (e.g., pairing physical robots with large language models), or new service features that use robots (e.g., a personal robot butler). It is also important to reward loyal customers, such as by offering exclusive interactions with high-end robots. In turn, their positive experiences can be leveraged for organic marketing for a firm's robot services – e.g., an 'Ambassador for Robots' program that offers benefits to customers in return for sharing their interactions and experiences with robots on social media.

Finally, our analyses revealed that irrespective of the degree of a robot's humanlike design, each attitude profile reflects a different threshold for perceiving a service robot as friendly or threatening. Such perceptions are important determinants of how customers respond to various service situations – for example, in service failure situations, the perceived lack of friendliness in a humanoid robot leads to greater dissatisfaction (Choi, Mattila, and Bolton 2021). Considering anthropomorphic effects (i.e., perceived sociability, uncanniness, and agency) in any attitude profile, Nimbro (ABOT humanlikeness score: 48.2) was the most preferred, while Nadine (ABOT humanlikeness score: 96.95) was the least preferred. Thus, a robot with median humanlikeness (scores range from 0 to 100 as per the ABOT database) would appeal to the broadest range of customer attitudes.

**Limitations and Further Research**

We investigated contextual and anthropomorphic effects using lab experiments, and future research may replicate these with actual interactions for additional validity. Further research can also examine how the attitude profiles influence behavioral patterns in various other service contexts. For example, research could compare service contexts that differ in terms of high/low customer involvement or risk – e.g., child-minding robot vs. coffee-making robot. Relatedly, the notion of robots as work colleagues and collaborative partners is fast becoming reality. As such, it is worth investigating any differences among attitude profiles in terms of interactions with robots in work teams and in various occupational contexts. From a servicescape perspective, researchers may wish to explore if/how changes to service design affect the link between attitude profiles and customer evaluations or satisfaction. For instance, how can a service situation be designed specifically to appeal to particular attitude profiles (e.g., ambivalent or indifferent)?

The implications of the attitudinal profiles may also be extended to service scenarios that are beyond conventional customer interactions with robots. For example, emerging research has highlighted the potential for human-AI service co-production, where service employees collaborate with robots to serve customers (Blaurock, Büttgen, and Schepers 2024; Le et al. 2024). Further studies could examine how the effectiveness of such collaborations may vary based on service robots paired with employees who exhibit characteristics of the four different attitude profiles uncovered by our research. Thus, as the field advances, our research may serve as a foundation for better understanding heterogeneous reactions to service robots.